\documentclass{article}

\usepackage{microtype}
\usepackage{graphicx}
\usepackage{subfigure}
\usepackage{booktabs} 
\usepackage{amsmath}
\usepackage{makecell}
\usepackage{color}
\usepackage{pifont}
\usepackage{hyperref}

\usepackage[accepted]{mlsys2024}

\mlsystitlerunning{Rethinking Memory and Communication Costs for Efficient Large Language Model Training}

\begin{document}

\twocolumn[
\mlsystitle{Rethinking Memory and Communication Costs for Efficient Large Language Model Training}



\mlsyssetsymbol{equal}{*}

\begin{mlsysauthorlist}
\mlsysauthor{Chan Wu}{ant}
\mlsysauthor{Hanxiao Zhang}{ant}
\mlsysauthor{Lin Ju}{ant}
\mlsysauthor{Jinjing Huang}{ant}
\mlsysauthor{Youshao Xiao}{ant}
\mlsysauthor{Zhaoxin Huan}{ant}
\mlsysauthor{Siyuan Li}{ant}
\mlsysauthor{Fanzhuang Meng}{ant}
\mlsysauthor{Lei Liang}{ant}
\mlsysauthor{Xiaolu Zhang}{ant}
\mlsysauthor{Jun Zhou}{ant}
\end{mlsysauthorlist}

\mlsysaffiliation{ant}{Ant Group, Beijing, Chaoyang, China}

\mlsyscorrespondingauthor{Chan Wu}{wuchan.wu@antgroup.com}
\mlsyscorrespondingauthor{Lin Ju}{julin.jl@antgroup.com}

\mlsyskeywords{Large Language Model, Partial Redundancy Optimizer, Training, Hierarchical Overlapping Ring, Communication Topology}

\vskip 0.3in

\begin{abstract}
Recently, various distributed strategies for large language model training have been proposed.
However, these methods provided limited solutions for the trade-off between memory consumption and communication cost.
In this paper, we rethink the impact of memory consumption and communication costs on the training speed of large language models, and propose a memory-communication balanced strategy set \underline{Pa}rtial \underline{R}edundancy \underline{O}ptimizer (PaRO).
PaRO provides comprehensive options which reduces the amount and frequency of inter-group communication with minor memory redundancy by fine-grained sharding strategy, thereby improving the training efficiency in various training scenarios.
Additionally, we propose a Hierarchical Overlapping Ring (HO-Ring) communication topology to enhance communication efficiency between nodes or across switches in large language model training. 
Our experiments demonstrate that PaRO significantly improves training throughput by 1.19$\times$-2.50$\times$ compared to the SOTA method and achieves a near-linear scalability.
The HO-Ring algorithm improves communication efficiency by 36.5\% compared to the traditional Ring algorithm.
\end{abstract}
]




\section{Introduction}
With the development of machine learning technology, the overall performance of deep learning algorithms in fields such as face recognition, recommender system, and natural language processing has significantly improved \cite{girshick2014rich,G-Meta,NEURIPS2020_1457c0d6}. 
Recent research shows that large model training is beneficial to improve model quality. 
Over the past few years, model size has increased from 110 million parameters for BERT \cite{devlin2019bert} to 175 billion parameters for GPT-3 \cite{NEURIPS2020_1457c0d6}.
However, training such large language model (LLM) is not an easy task, as it requires a significant amount of computing resources and presents challenges in terms of system complexity.

As the size of the model and the amount of training data increase, the computing power of a single GPU cannot meet the training needs of large-scale networks. 
In LLM training, to effectively utilize the computing power and memory of hundreds of GPU devices, a variety of distributed parallel training technologies have been proposed, such as data parallelism (DP), tensor parallelism (TP) and pipeline parallelism (PP) \cite{2023ColossalAI}.
In DP, an entire dataset is evenly partitioned into mutually exclusive subsets before training, and each worker works on a separate subset of them. 
TP divides the calculation and memory load of a single layer onto multiple GPUs by modifying the calculation method within the layer. 
PP puts different layers on different GPUs, and then divides the computing and memory loads onto multiple GPUs. 
However, TP and PP require modification of the model implement, which is inefficient for developers.
In contrast, data parallelism has become the most mainstream distributed parallel method due to its simplicity.

In data parallelism, the replicated model on each GPU processes a portion of the input batch, resulting in a large amount of communication data when fusing gradients.
Andrew \cite{baidu-allreduce} applied a ring topology on all-reduce to balance the communication load.
By defining the communication topology, the communication pressure is evenly distributed to each GPU.
However, since a complete model is copied on each GPU, significant memory redundancy occurs, especially when training large models \cite{proficz_improving_2018}.
To this end, Rajbhandari et al. \cite{ZeRO} proposed the Zero Redundancy Optimization (ZeRO) strategy set, which splits the model state (i.e. optimizer state, gradient and parameters) based on data parallelism and reconstructs them through the collective communication.
It reduces memory consumption in LLM training and improves training efficiency by applying larger batch sizes.

Since ZeRO retains the simplicity, ease of use, and versatility of DP, it has been widely used in LLM training.
ZeRO needs to be adapted to specific training frameworks and hardware equipment to fully exploit its advantages.
In high-performance clusters such as NVIDIA DGX-2 or DGX-A100 \cite{Blink2020}, NVLink/NVSwitch with a bandwidth of up to 4.8TGbps is configured within the node, while the bandwidth of InfiniBand or Ethernet between nodes is only 200$\sim$800Gbps.
The mismatch of bandwidth within and between nodes limits the training efficiency of ZeRO.
To speed up model training, ZeRO requires more GPU resources, which will result in greater collective communication volume.
To reduce collective communication volume, MiCS \cite{MiCS22} proposes a cluster grouping strategy in which all model states are partitioned within each group and replicated across different groups. 
However, this partitioning strategy incurs significant memory costs, particularly in scenarios with a large number of groups.

In this paper, we systematically combine cluster grouping with different partitioning of different model states to trade off the memory and communication costs.
Based on the memory consumption and synchronization frequency of the optimizer state, gradients and parameters, we design several optimization solutions to reduce overall communication costs and frequency with minimal memory redundancy. 
Additionally, we optimize the communication topology of ring all-gather and reduce-scatter operations by performing intra- and inter-node communication simultaneously. 
This strategy reduces inter-node communication volume and improves inter-node bandwidth utilization. 
We plan to release the code, pending approval from the company.
The main contributions of the paper are summarized as follows:
{
\setlength{\parskip}{0pt}
\begin{itemize}
    \setlength{\parskip}{0pt}
    \item We systematically analyzed the impact of memory consumption and communication costs on the training speed of LLMs, and proposed an overall guideline for balancing memory and communication.
    \item We proposed the Partial Redundancy Optimizer (PaRO) strategy set, which provides more refined options for the trade-off between memory consumption and communication costs in different training scenarios.
    PaRO significantly increased training throughput by 1.19$\times$-2.50$\times$ comparing with ZeRO, and can also improve the efficiency of complex ML systems.
    \item We proposed a Hierarchical Overlapping Ring (HO-Ring) communication topology for inter-node or cross-switch collective communication operations for LLM training or other scenarios. 
    Compared with the traditional Ring, the communication efficiency of HO-Ring was increased by 36.5\%.
\end{itemize}
}

\section{Background and Related Works}
\subsection{Data and Tensor Parallelism}
According to different parallel objects, distributed parallel training technology can be divided into data parallelism and tensor parallelism \cite{2023reducing}.

Data parallelism divides the input data equally into several shards and assigns them to different GPUs.
Each GPU owns the complete replica of model parameters.
After forward and backward computation, each GPU obtains the corresponding parameter gradients. 
These gradients are then aggregated and transmitted back to each GPU through the all-reduce operation. 
Finally, the model parameters are updated based on the gradient and optimizer state \cite{sergeev2018horovod}. 
Data parallelism simplifies model training and deployment, but requires each GPU to maintain a complete replica of the model state. 
It may not meet the memory requirements of LLMs, especially when using the Adam optimizer \cite{kingma2017adam}. 
Additionally, the communication cost during gradient transmission increases almost linearly with the number of GPUs, making the network bandwidth a bottleneck for training efficiency.

Tensor parallelism shards tensors onto multiple GPU devices by modifying the model structure, and implements model parallelism through distributed matrix multiplication.
Based on the characteristics of the Transformer architecture, Megatron-LM \cite{2019Megatron} divides the layers in the row or column dimension to achieve 1D tensor parallelism.
Since the output of each layer in 1D tensor parallelism is incomplete, an all-gather operation is required to aggregate the complete input before passing it to the next layer.
In this process, the collective communication of 1D tensor parallelism generates a large amount of communication cost.
Low bandwidth between nodes will affect the efficiency of 1D tensor parallel training.
Additionally, 1D tensor parallelism incurs redundant memory consumption due to repeated inputs to each layer and repeated outputs after all-reduce.
To address these issues, more advanced tensor parallelism methods, such as 2D \cite{2DMethod2023}, 2.5D \cite{wang2021}, and 3D \cite{bian2021maximizing} tensor parallelism, have been introduced in LLM training. 
These methods shard the initial inputs using distributed matrix multiplication \cite{2_5D2011, 5389455}, which eliminates communication in the middle layer and only requires one all-gather communication in the last layer.


\subsection{Model Finetuning}
Powerful BERT\cite{devlin2019bert} and GPT3\cite{NEURIPS2020_1457c0d6} models are both pre-trained on a large amount of general domain data. A widely-used approach, fine-tuning, freezes part of the pre-trained parameters and finetunes the remaining layers on task-specific data provides a significant performance and efficiency gain in different domains \cite{girshick2014rich,NEURIPS2020_1457c0d6}. Different fine-tuning approaches vary on the ratio of trainable parameters of existing pre-trained models, including full parameter finetuning, and partial parameter finetuning\cite{lialin2023scaling}. The full parameter finetuning is as expensive as the pre-training since all model states must be stored. In the partial parameter finetuing, only the parameters are required to fully stored for computation, while the gradients and optimizer states are limited to trainable parameters. However, the enormity of pre-trained model, such as GPT-3, makes it challenging to perform traditional partial fine-tuning, so Parameter-efficient fine-tuning (PEFT), such as LoRA\cite{hu2022lora}, P-tuning\cite{P-Tuning}, is introduced to resolve this problem by only training a very small set of parameters, which might be a subset of the existing model parameters or a set of newly added parameters.

\subsection{ZeRO Optimizer}
The training process of deep learning models mainly consists of three stages: forward computation, backward computation, and model update.
During the training process, GPUs need to store both model state and residual memory. ZeRO \cite{ZeRO} primarily reduces the memory consumption of model states, which mainly include model parameters, gradients from backward computation, and optimizer states for parameter updates. ZeRO gradually optimizes redundant memory in three stages: ZeRO-1, ZeRO-2 and ZeRO-3.

ZeRO-1 globally shards the optimizer state across all GPU devices. During the training process, each GPU performs forward and backward computation independently to obtain the gradient, which are then synchronized among all GPUs using the all-reduce operation.
Since each GPU retains a shard of the optimizer state, only the corresponding model parameters can be updated.
After that, the updated model parameter shards are retrieved from other GPUs using the all-gather operation to ensure that all GPUs have the latest model parameters.

Compared to ZeRO-1, ZeRO-2 further shards the optimizer state.
During the training process, each GPU stores a complete set of model parameters and independently performs forward and backward computation to obtain a gradient.
Afterwards, each GPU updates the gradient shards through the reduce-scatter operation and discards the other gradient shards. The subsequent processes remain the same as in ZeRO-1.

In ZeRO-3, model parameters, gradients, and optimizer state are all sharded.
Before performing forward and backward computation, each GPU performs an all-gather operation to collect model parameter shards from other GPUs and construct the complete model parameters. 
After gradient calculation, each GPU immediately discards the unmaintained model parameter shards.
Then, each GPU updates the corresponding shard of model parameters using the maintained shard of optimizer parameters and gradients.
Since each GPU only maintains one model parameter shard, there is no need to perform all-reduce operations.


\subsection{Communication Cost} \label{sec:background_comm_cost}
For models with billions to trillions of parameters, ZeRO-3 transfers a significant amount of data ranging from tens to hundreds of gigabytes during forward computation, backward computation, and model updates.
As the cluster size grows, each GPU needs to communicate multiple times, which amplifies the latency of collective communication operations.
Therefore, an efficient communication topology is crucial to reduce communication costs.

For communication primitives, the traditional ring all-reduce fails to consider the differences in intra- and inter-node network bandwidth, thereby being unable to fully utilize the bandwidth of clusters.
The hierarchical ring (H-Ring) all-reduce \cite{jia2018highly} groups GPUs based on their respective nodes and improves the efficiency through the communication topology of intra-group reduce, inter-group all-reduce, and intra-group broadcast.
However, in inter-group all-reduce, only one GPU of each node participates in communication, resulting in low inter-group bandwidth utilization.
To address this issue, Mikami et al. \cite{Mikami2018MassivelyDS} proposed the 2D-Torus all-reduce scheme, where the communication topology is modified into intra-group reduce-scatter, inter-group all-reduce and intra-group all-gather.
While the total communication volume of 2D-Torus all-reduce is the same as H-Ring all-reduce, 2D-Torus is more efficient due to the simultaneous communication of all GPUs in inter-group all-reduce.

The community further optimizes the communication cost based on the inherent characteristic of LLM. 
To reduce the inter-node communication costs, MiCS \cite{MiCS22} introduces the group sharding strategy by dividing the GPU cluster into subgroups, where the model state is partitioned within the subgroups and replicated across the subgroups.
By configuring suitable subgroup sizes, MiCS can leverage the high intra-node bandwidth and a hierarchical communication strategy to reduce the communication volume between nodes.
Similarly, the ZeRO++ \cite{wang2023zero++} system performs a secondary sharding of parameters while keeping other model states sharded across all GPUs to reduce inter-node communication volume.
In addition, ZeRO++ compresses model parameters and gradients through quantization to reduce inter-node communication volume and latency.
Additionally, the PyTorch's official Fully Sharded Data Parallel \cite{FSDP23} provides a hybrid sharding (FSDP-hs) strategy, which leverages data center locality to accelerate training and reduce inter-node communication.

\section{PaRO Design}
\subsection{Analysis and Insights in LLM Training}
This subsection examines the memory and communication costs of LLM training using the group sharding strategy. 
We consider training tasks with different model sizes and three levels of trainable model parameters: full, partial, and PEFT. We refer finetuning as partial training tasks for simplicity and introduce the following notations to aid in the explanation:
{
\setlength{\parskip}{0pt}
\begin{itemize}
    \setlength{\parskip}{-2pt}
    \item []   $N$: Number of GPUs in the cluster.
    \item []   $M$: Number of GPUs in the group or node.
    \item []   $g$: Number of groups or nodes, $g=N/M$.
    \item []   $s$: Step of gradient accumulations.
    \item []   $K$: Optimizer parameters.
    \item []   $\Psi$: Number of model parameters.
    \item []   $\Psi'$: Number of trainable parameters.
    \item []   $P$: Parameter.
    \item []   $G$: Gradient.
    \item []   $OS$: Optimizer state.
\end{itemize}
}


\subsubsection{Analysis of Communication Cost}
As mentioned in the subsection ~\ref{sec:background_comm_cost}, there exists a substantial performance gap in the bandwidth and latency between intra- and inter-node networks, which bottlenecks the training efficiency. Grouping GPU with a little memory redundancy can reduce communication participants and communication costs. 
Additionally, the subgroup could be grouped within the intra- and inter-node networks to fully leverage the high-throughput intra-node network.
It could significantly improve communication efficiency.

Therefore, we define three sharding states: \emph{no sharding}, \emph{intra-group sharding} and \emph{global sharding}, based on the sharding scope for three components of model states. 
\underline{The order of sharding granularity}, from coarse-grained to fine-grained, is as follows: \emph{no sharding} $>$ \emph{intra-group sharding} $>$ \emph{global sharding}. 
More specifically, \emph{intra-group sharding} means that model states are sharded within the group, while each group holds the complete replica. 
\emph{No sharding} means that each GPU holds a replica of model states, while each GPU holds a part of model states in \emph{global sharding}. 

In the context of gradient accumulation where one mini-batch step contains several mirco-batch steps, we analyze the communication cost of model states with different sharding states.
{
\setlength{\parskip}{0pt}
\begin{itemize}
    \setlength{\parskip}{-2pt}
    \item \textbf{Parameter sharding:} 
    Parameters are utilized in both forward and backward computations during each iteration of micro-batch. In the both \emph{global sharding} and \emph{intra-group sharding} states, an all-gather operation is necessary to obtain all parameters of the current layer before usage. While only intra-group all-gather is required when sharding model parameters within a group. It reduces the frequency of high time-cost inter-group communication with little redundant memory across the inter-group. In \emph{no sharding} state, no communication operation is required since each device holds replicated parameters.
    
    \item \textbf{Gradient sharding:} 
    Gradients are computed during the backward computation and used in the model update stage. Likewise, in both \emph{global sharding} and \emph{intra-group sharding} states, the aggregated gradient of the corresponding local shard is obtained through collective communication. When sharding gradients within a group, only intra-group reduce-scatter is required in mini-batch step. After a number of gradient accumulations in a mini-batch step, an intra-group or global reduce-scatter operation is performed depending on the sharding scope of the optimizer state. In \emph{no sharding} state, no communication operation is required before parameter update stage.
    
    \item \textbf{Optimizer state sharding:} 
    The optimizer state is utilized during the model updating stage. However, the communication operations before or after the model updating stage become more complicated since they depend on the consistency of sharding scope between gradients and model parameters. If the sharding scope of optimizer states differs from that of gradients or parameters, the communication operations will vary before and after the model updating stage. For instance, it requires to perform an inter-group reduce-scatter before model updating and an inter-group all-gather after model updating, when the OS is global sharding and others are intra-group sharding. 
    
\end{itemize}
}
    


Therefore, \underline{the order of communication costs} is as follows:
\emph{no sharding} $<$ \emph{intra-group sharding} $<$ \emph{global sharding}. Additionally, we highlight that the communication bottleneck vary with trainable parameters.
{
\setlength{\parskip}{0pt}
\begin{itemize}
    \setlength{\parskip}{-2pt}
    \item When $\Psi'\le\Psi$ in the full or partial parameters training, the bottleneck lies in the inter-node communication bandwidth.
    \item When $\Psi'\ll\Psi$, e.g. PEFT, the bottleneck lies in the communication frequency. This is because a large amount of fragmented communications reduces overall bandwidth utilization. 
\end{itemize}
}

\subsubsection{Analysis of Memory Cost}
In contrast to ZeRO optimizer, we account for the memory consumption of the model states with an extra \emph{intra-group sharding} state. 
Obviously, \underline{the order of memory consumption} is: \emph{global sharding} $<$ \emph{intra-group sharding} $<$ \emph{no sharding}, which is inverse to the order of communication cost. 
Memory savings come at the cost of increased communication. 
In the mainstream mixed precision training using Adam optimizer\cite{kingma2017adam}, the memory consumption of the parameters, gradients and optimizer states are respectively $2\Psi$, $2\Psi'$, and $12\Psi'$. 
In PEFT tasks, the sizes of G and OS are relatively small compared to P of mega pre-trained models. We summarize:
{
\setlength{\parskip}{0pt}
\begin{itemize}
    \setlength{\parskip}{-2pt}
    \item When $\Psi'\le\Psi$, optimizer states consume the most memory.
    
    \item When $\Psi' \ll \Psi $, parameters consumes the most memory, followed by optimizer states and model parameters.
    
\end{itemize}
}

\subsubsection{Trade-off between Memory and Communication}
The above three levels of sharding granularity on P, G and OS brings up 27 combinations of model sharding strategies. 
Generally, utilizing a more fine-grained sharding level can save memory for larger batch input needs, and thus increases throughput per GPU. 

Therefore, there exists a trade-off between memory savings and communication costs when selecting the appropriate model sharding strategy. In all scenarios, any strategy with sharding priority $S_{OS} > S_P$ or $S_{OS} > S_G$ is inferior to the corresponding strategy with $S_{OS} = S_P$ or $S_{OS} = S_G$. This is because the former not only consumes more memory but also fails to achieve any savings in communication overhead compared to the latter. Based on this key insight, it infers to \textbf{Principle 1} that $S_P \geq S_{OS}$ and $S_G \geq S_{OS}$ in terms of the order of sharding granularity for all levels of trainable parameters. In other words, a more fine-grained shard strategy should be employed for OS compared to P and G. According to this principle, we can eliminate 13 out of the 27 possible combinations mentioned earlier.

Secondly, when $\Psi'\ge \frac{\Psi}{6}$, the memory consumption of $P$ is greater than or equal to that of $G$. In the full parameter training when $\Psi' = \Psi$, both $P$ and $G$ own the same memory consumption, however, the communication frequency of $P$ is as twice as $G$. 
This is because $P$ is utilized in both the forward pass and backward pass, while $G$ is only used in the backward propagation.
Furthermore, in the partial parameter training when $\Psi' > \frac{\Psi}{6}$, the memory consumption of gradients reduce to $\Psi'$ while the $P$ is still $\Psi$ since all of the parameters have to be utilized in the training. 
Therefore, we infer that $S_P \geq S_G$. Combined with Principle 1, we achieve \textbf{Principle 2} that $S_P \geq S_G \geq S_{OS}$ when $\Psi'\ge \frac{\Psi}{6}$.

Thirdly, in PEFT training tasks when $\Psi' \ll \Psi$, the memory consumption of $P$ is still $\Psi$ to be used in the forward computation while $G$ and $OS$ are quite small. In this case, sharding $G$ would result in a negligible amount of memory savings but would lead to increased communication overhead. This infers to \textbf{Principle 3} that $G$ should not be sharded in PEFT training.

\subsection{Partial Redundancy Optimizer}

\begin{table}[t]
    \caption{Optional sharding strategies for varying numbers of trainable parameters. P/G/OS represents the combination of sharding strategies for Parameter/Gradient/Optimizer, N: no sharding, I: intra-group sharding, G: global sharding. $\Psi'=\Psi$, $\Psi'\ge\frac{\Psi}{6}$ , $\Psi'<\frac{\Psi}{6}$ and PEFT means the different ratios of trainable parameters to model parameters. \ding{51} is recommended while \ding{55} is the opposite. }
    \label{tab:PaROs}
    \begin{center}
    \begin{small}
    \begin{tabular}{lcccc}
    \toprule
    P/G/OS & $\Psi'=\Psi$ & $\Psi'\ge\frac{\Psi}{6}$ & $\Psi'<\frac{\Psi}{6}$ & PEFT \\
    \midrule
        \texttt{NNN(DDP)} & \ding{51} & \ding{51} & \ding{51} & \ding{51}  \\
        \texttt{NNI} & \ding{51} & \ding{51} & \ding{51} & \ding{51}  \\
        \texttt{NNG(ZeRO-1)} & \ding{51} & \ding{51} & \ding{51} & \ding{55}  \\
        \texttt{NII} & \ding{51} & \ding{51} & \ding{51} & \ding{55}  \\
        \texttt{NIG} & \ding{51} & \ding{51} & \ding{51} & \ding{55}  \\
        \texttt{NGG(ZeRO-2)} & \ding{51} & \ding{51} & \ding{51} & \ding{55}  \\
        \texttt{INI} & \ding{55} & \ding{55} & \ding{55} & \ding{51}  \\
        \texttt{ING} & \ding{55} & \ding{51} & \ding{51} & \ding{55}  \\
        \texttt{III(MiCS)} & \ding{55} & \ding{55} & \ding{51} & \ding{55}  \\
        \texttt{IIG} & \ding{51} & \ding{51} & \ding{55} & \ding{55}  \\
        \texttt{IGG} & \ding{51} & \ding{51} & \ding{51} & \ding{55}  \\
        \texttt{GNG} & \ding{55} & \ding{51} & \ding{51} & \ding{51}  \\
        \texttt{GIG} & \ding{55} & \ding{51} & \ding{51} & \ding{55}  \\
        \texttt{GGG(ZeRO-3)} & \ding{51} & \ding{51} & \ding{51} & \ding{55}  \\
    \bottomrule
    \end{tabular}
    \end{small}
    \end{center}
\end{table}


Although ZeRO and MiCS advanced the development of LLM training, they only provide limited solutions. Based on Principle 1, we can filter out 14 meaningful combinations from the previously mentioned 27 combinations. These 14 combinations form our proposed PaRO strategy set, which is presented in Table 1. Among these solutions, DDP, ZeRO and MiCS can be regard as special cases within the PaRO.
Based on Principles 2 and 3, we can deduce that certain strategies are meaningless under the conditions of $\Psi' \geq \frac{\Psi}{6}$ and PEFT.
Additionally, based on specific scenarios, more meaningless strategies can be eliminated.For example, in the case of $\Psi' = \frac{\Psi}{6}$, PaRO-INI is always inferior to PaRO-IIG, due to both the memory and communication costs of the former are greater than those of the latter.
Another example is that, in the case of PEFT, PaRO-ING is always worse than PaRO-INI. This is because, for PEFT training, the memory consumption is almost identical for both strategies, but the former has higher communication overhead than the latter.
Furthermore, we argue that the comprehensive PaRO strategy set provides more flexibility to complicated machine learning system, such as distributed RLHF system \cite{ouyang2022training}. 
In the following paragraphs, we provide a detailed explanation of three PaRO solutions as running examples in full parameter training: PaRO-IGG, PaRO-IIG, and PaRO-NIG. The implementation of other PaRO strategies can be easily derived from these three solutions.

\begin{figure}[t]
    \centering
    \includegraphics[width=0.48\textwidth]{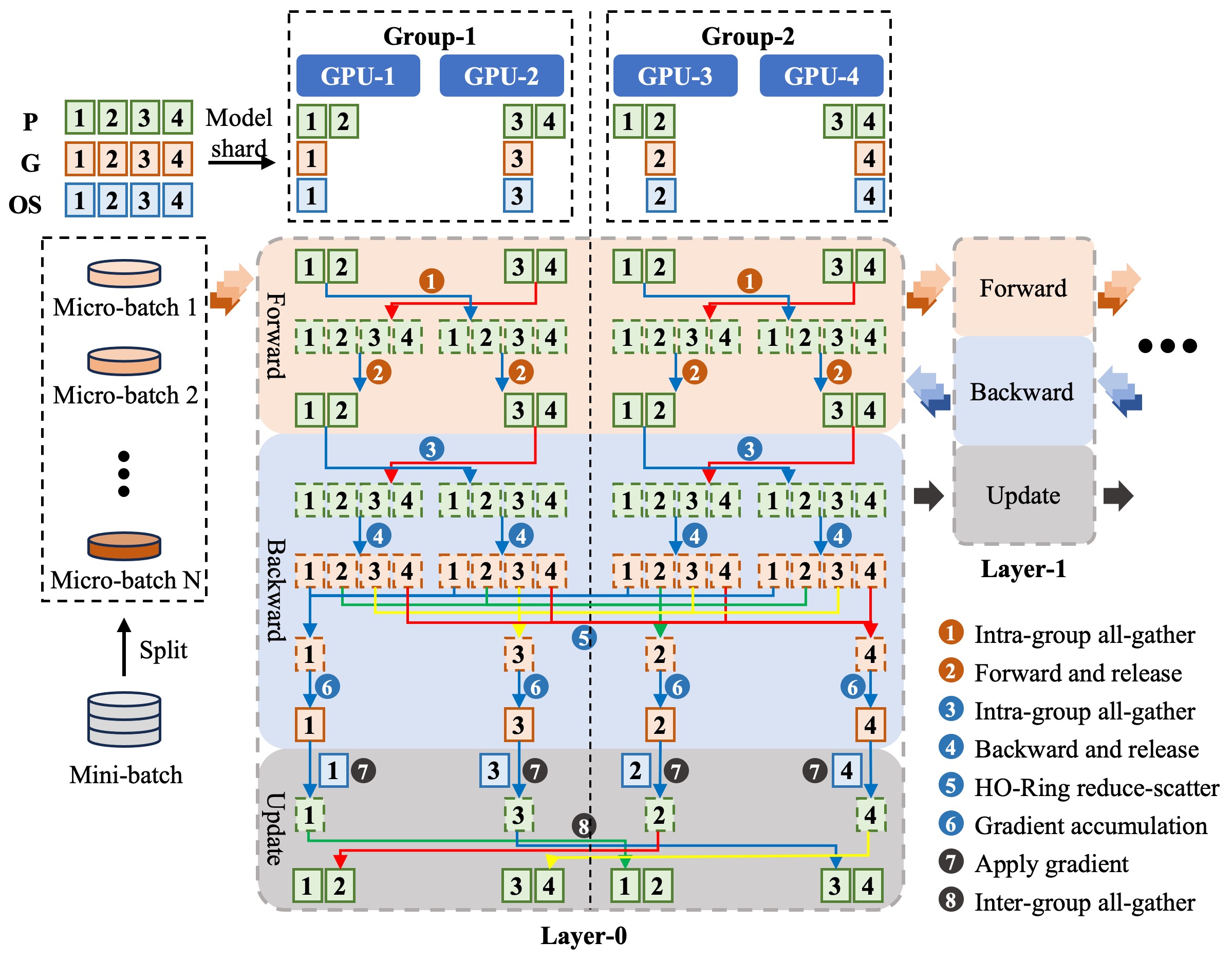}
    \caption{Schematic of PaRO-IGG in a grouped cluster with four GPUs. 
    The parameters (P) of the model are sharded within the group, while gradients (G) and optimizer states (OS) are sharded globally. 
    Labeled rectangular blocks represent shards of model parameters, gradients, and optimizer states. 
    The solid and dashed rectangular blocks represent fixed and temporary shards respectively. 
    Circular nodes represent operations in different stages (Forward, Backward, and Update). 
    After feeding a micro-batch, the model will sequentially execute the Forward stage of each layer, followed by the reverse execution of the Backward stage of each layer. 
    The Update stage will only be executed after completing the Backward stage of the last micro-batch.}
    \label{fig:PaRO-IGG}
\end{figure}

\begin{figure}[t]
    \centering
    \includegraphics[width=0.48\textwidth]{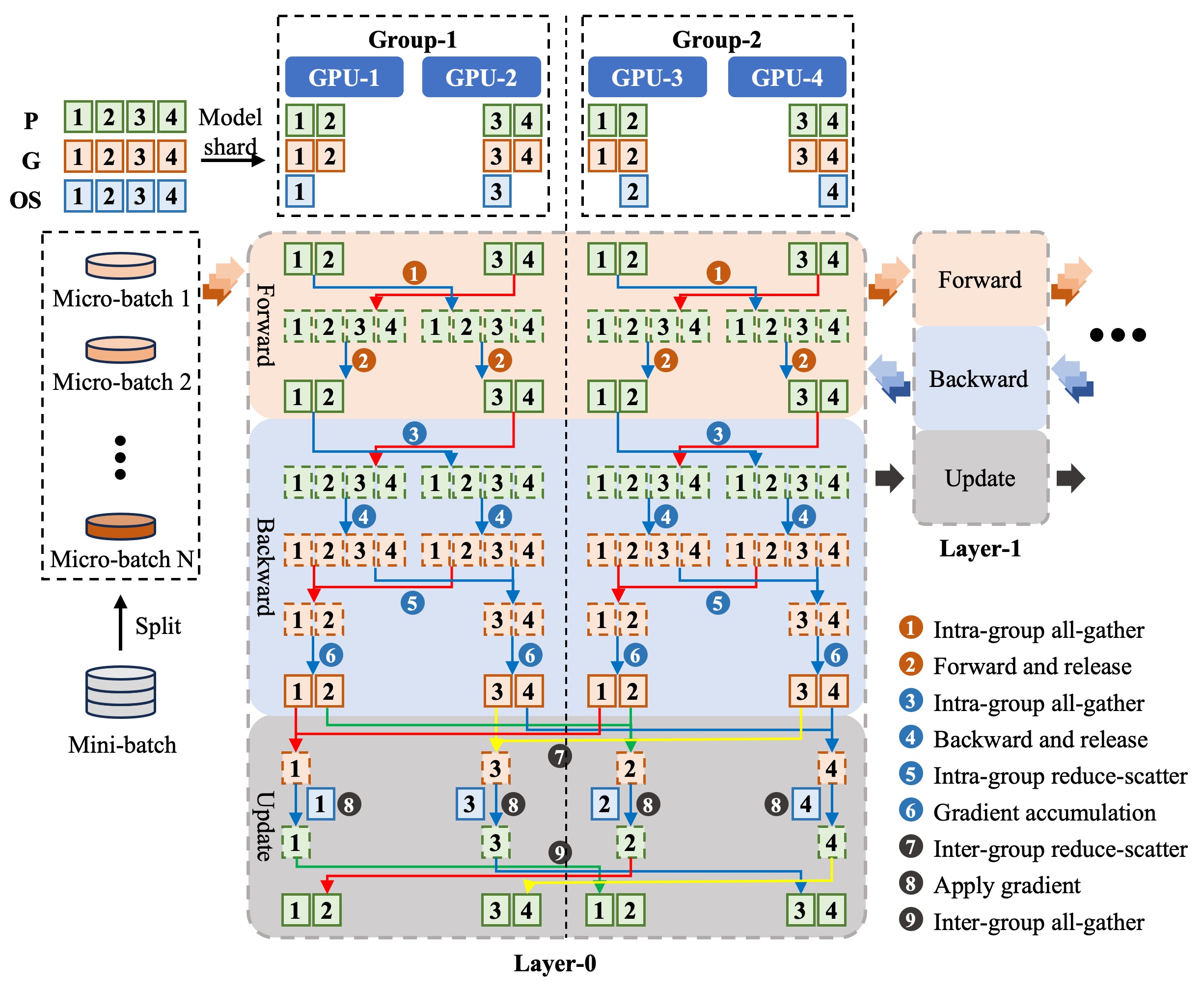}
    \caption{Schematic of PaRO-IIG in a grouped cluster with four GPUs.}
    \label{fig:PaRO-IIG}
\end{figure}

\begin{figure}[t]
    \centering
    \includegraphics[width=0.48\textwidth]{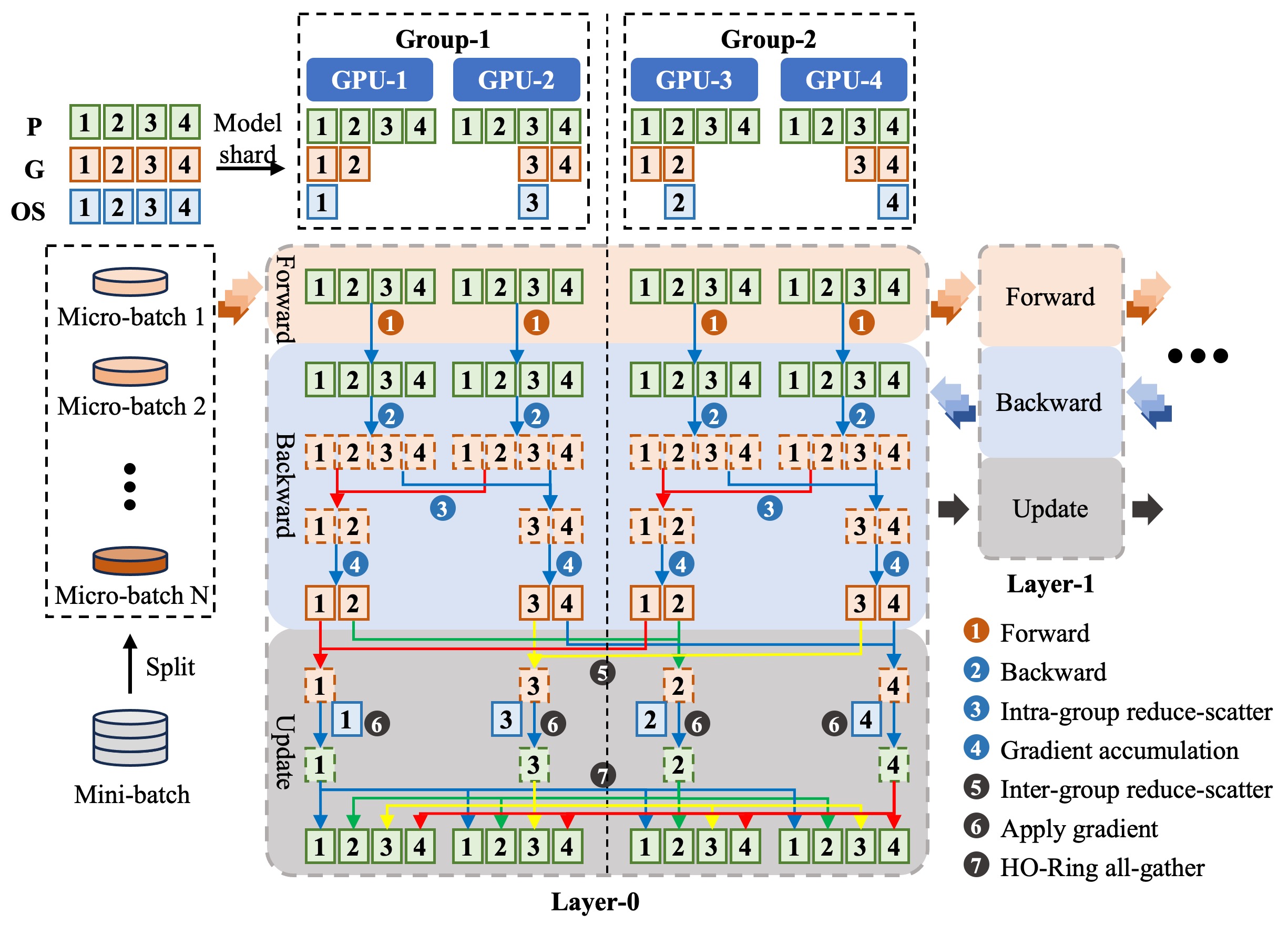}
    \caption{Schematic of PaRO-NIG in a grouped cluster with four GPUs.}
    \label{fig:PaRO-NIG}
\end{figure}

Figure ~\ref{fig:PaRO-IGG} illustrates the schematic of PaRO-IGG.
To simplify the diagram, we only use four GPUs and divide them into two groups.
To reduce inter-group communication frequency and volume, model parameters are intra-group sharded, while gradients and optimizer states are globally sharded.
Therefore, a complete replica of the model parameters is stored within each group.
During the training process, a mini-batch is divided into multiple micro-batches to reduce the memory consumption for storing activation outputs.
In the Forward stage, each GPU obtains a complete replica of model parameters through the intra-group all-gather operation.
These model parameters are used to perform the forward computation of the current layer on the input micro-batch, and are later released to reduce GPU memory consumption.
After completing the Forward stage of the current layer, the system proceeds to the Forward stage of the next layer until the final layer of the network.
In the Backward phase, the model parameters are collected again through the intra-group all-gather and released after the backward computation of this layer. 
After the backward computation, each GPU obtains a complete replica of the gradients and releases the redundant model parameters.
Each GPU aggregates gradients from other GPUs through HO-Ring reduce-scatter operations for global gradient synchronization.
In addition, each GPU maintains a gradient shard that accumulates gradients generated by each micro-batch.
Similarly, after completing the Backward phase of the current layer, the system will execute the Backward phase of the previous layer until the first layer of the network.
Once the gradients of the last micro-batch are accumulated, each GPU utilizes the gradient shard to update the optimizer state maintained by itself and generate low-precision model parameters.
Finally, model parameter shards are obtained from other groups through an inter-group all-gather operation.

Figure ~\ref{fig:PaRO-IIG} illustrates the schematic of PaRO-IIG.
Different from PaRO-IGG, in PaRO-IIG, the model parameters and gradients are intra-group sharded, while the optimizer states are globally sharded.
Therefore, full model parameters and gradients are preserved within each group.
In the Forward and pre-Backward stages, the computation processes of PaRO-IIG and PaRO-IGG are consistent.
After the backward computation, each GPU aggregates gradients from other GPUs through intra-group reduce-scatter operations for local gradient synchronization.
These gradients are temporarily stored on each GPU through gradient accumulation.
Once the gradients of the last micro-batch are accumulated, each GPU performs an inter-group reduce-scatter operation to achieve global gradient synchronization.
The subsequent Update operations are the same as PaRO-IGG.

Figure ~\ref{fig:PaRO-NIG} illustrates the schematic of PaRO-NIG.
In PaRO-NIG, the parameters of the model are not sharded, the gradients are intra-group sharded, and the optimizer states are global sharded.
Different from the above two solutions, each GPU retains complete model parameters in PaRO-NIG.
Therefore, in the Forward and Backward stages, each GPU can directly perform the forward and backward computation without collecting and releasing model parameters.
The subsequent four-step computation process of PaRO-NIG is consistent with that of PaRO-IIG. 
Finally, each GPU collects updated gradients by HO-Ring all-gather operations.

\subsection{PaRO with Gradient Accumulation}
In PaRO, we introduce the gradient accumulation strategy \cite{li20211bit, you2020large} to obtain large batches of inputs.
Furthermore, we narrow the scope of gradient synchronization to reduce communication volume and frequency.
Specifically, we perform intra-group sharding and inter-group replication of gradients.
The gradients of each micro-batch are synchronized through the intra-group reduce-scatter.
After accumulating the gradients from all micro-batches, global gradient synchronization can be achieved by performing an inter-group reduce-scatter operation only once.
Compared with the global reduce-scatter, the single-GPU communication volume reduced by the grouped two-step reduce-scatter is calculated as follows:
\begin{equation} \label{eq3}
    \begin{split}
    \Delta_C&=s*\frac{\Psi}{N}*(N-1)-\\
    &\quad\Big(s*\frac{\Psi}{M}*(M-1)+\frac{\Psi}{N}*(g-1)\Big) \\
    &=\frac{\Psi*(s-1)*(g-1)}{N}
    \end{split}
\end{equation}
where, the first item is the communication volume of global reduce-scatter, and the second item is the total communication volume of intra- and inter-group reduce-scatters.
It can be observed that as the number of groups $g$ and the accumulation steps $s$ increase, the communication volume on a single GPU decreases further.
In the absence of cluster grouping (i.e. $g=1$) or gradient accumulation (i.e. $s=1$), there is no reduction in single-GPU communication volume.
Therefore, the combination of gradient accumulation and cluster grouping is of practical significance to reduce communication.

\subsection{HO-Ring for All-gather and Reduce-scatter}

\begin{figure}[t]
    \centering
    \includegraphics[width=0.48\textwidth]{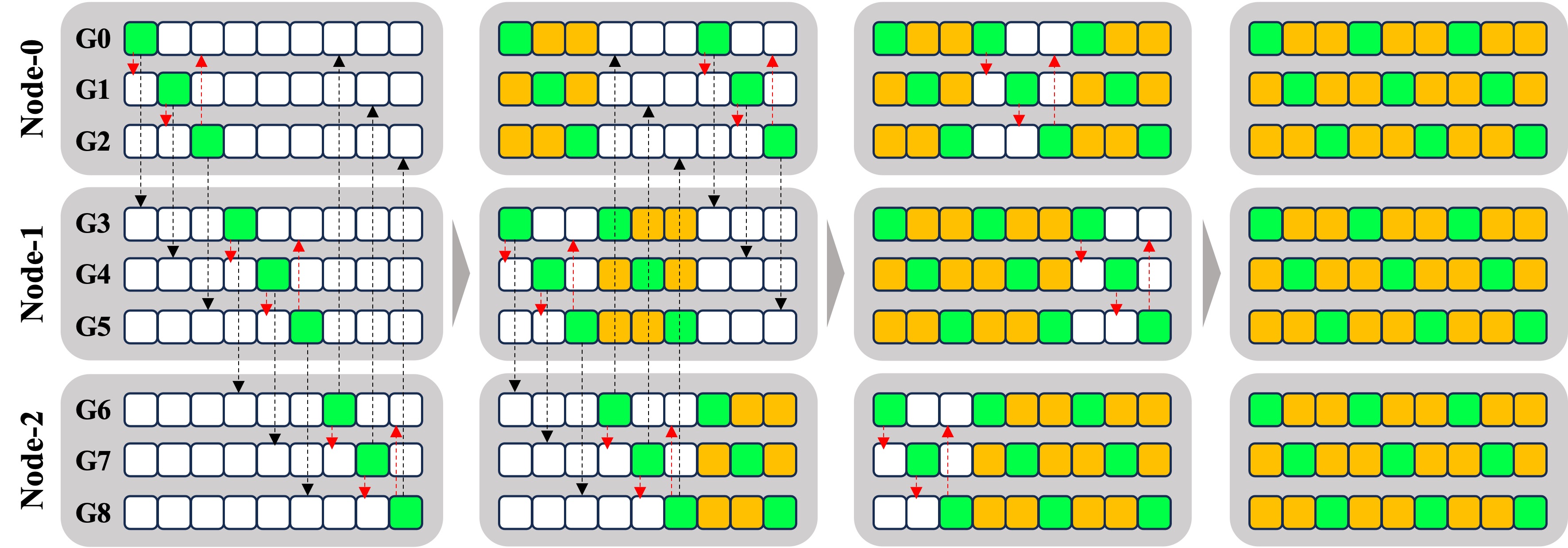}
    \caption{Communication topology of HO-Ring. The $N$ ($N=9$) GPUs (G0-G8) are divided equally into $g$ ($g=3$) groups.
        Red and black arrows represent intra- and inter-group communication respectively.
        Orange and green blocks represent data obtained through intra- and inter-group communication, respectively.
        }
    \label{fig:HO-Ring}
\end{figure}

Since the model state is sharded in a GPU cluster, it is necessary to aggregate or scatter these shards for global synchronization, 
such as the all-gather for parameters in PaRO-NIG and the reduce-scatter for gradients in PaRO-IGG.
In the traditional Ring, each GPU sequentially transfers its shard of data to the next GPU.
The transmission efficiency of cross-node communication may be a bottleneck affecting model training.
The H-Ring groups GPUs based on their respective nodes.
The global all-gather/reduce-scatter is divided into two steps: intra- and inter-group all-gather/reduce-scatter, to improve inter-group bandwidth utilization and avoid partial GPU waiting.
However, during inter-group communication, the intra-group bandwidth is idle, resulting in a waste of resources. 
Therefore, we proposed a HO-Ring communication topology for all-gather/reduce-scatter.

Figure ~\ref{fig:HO-Ring} shows the communication topology of HO-Ring. 
Like H-Ring, the GPUs in HO-Ring are also grouped based on their respective nodes.
Each GPU transmits its own shards simultaneously through the intra- and inter-group communication rings, as shown in the first two steps in Figure ~\ref{fig:HO-Ring}. 
Different from the H-Ring, HO-Ring can simultaneously utilize communication resources within and between groups to improve transmission efficiency.
After the inter-group communication ring is completed, an intra-group communication ring is executed to gather the remaining shards within the group, as shown in the third step in Figure ~\ref{fig:HO-Ring}.

\subsection{Memory and Communication Analysis}
This section analyzes the advantages of the above three solutions in terms of memory consumption and communication by comparing other solutions.

\begin{table}[t]
    \caption{Single-GPU memory consumption of parameter P, gradient G and optimizer state OS in different solutions.}
    \label{tab:mem}
    \vskip 0.15in
    \begin{center}
    \begin{small}
    \begin{tabular}{lccc}
    \toprule
    Model states & P & G & OS \\
    \midrule
    ZeRO-1       & $2\Psi$   & $2\Psi$               & $\frac{K\Psi}{N}$ \\\specialrule{0em}{0pt}{5pt}
    ZeRO-2       & $2\Psi$   & $\frac{2\Psi}{N}$     & $\frac{K\Psi}{N}$ \\\specialrule{0em}{0pt}{5pt}
    ZeRO-3       & $\frac{2\Psi}{N}$ & $\frac{2\Psi}{N}$ & $\frac{K\Psi}{N}$  \\\specialrule{0em}{0pt}{5pt}
    MiCS        & $\frac{2\Psi}{M}$ & $\frac{2\Psi}{M}$ & $\frac{K\Psi}{M}$  \\\specialrule{0em}{0pt}{5pt}
    ZeRO++      & $\frac{2\Psi}{N}+\frac{2\Psi}{M}$ & $\frac{2\Psi}{N}$  & $\frac{K\Psi}{N}$ \\\specialrule{0em}{0pt}{5pt}
    PaRO-IGG    & $\frac{2\Psi}{M}$ & $\frac{2\Psi}{N}$ & $\frac{K\Psi}{N}$ \\\specialrule{0em}{0pt}{5pt}
    PaRO-IIG    & $\frac{2\Psi}{M}$ & $\frac{2\Psi}{M}$ & $\frac{K\Psi}{N}$ \\\specialrule{0em}{0pt}{5pt}
    PaRO-NIG    & $2\Psi$   & $\frac{2\Psi}{M}$ & $\frac{K\Psi}{N}$ \\
    \bottomrule
    \end{tabular}
    \end{small}
    \end{center}
    \vskip -0.1in
\end{table}

\begin{table*}[t]
    \caption{Total communication volume of different solutions in the Forward, Backward and Update stages with a micro-batch input.
    A-G(P) represents the all-gather operation for the parameter P; 
    R-S(G) and \textbf{A-G(G)} respectively represent the reduce-scatter and all-reduce (\textbf{bold}) operations on the gradient G.
    The $^\dag$ symbol in the upper right corner of the data indicates that the operation is inter-group communication, otherwise it is intra-group communication.}
    \label{tab:com}
    \vskip 0.15in
    \begin{center}
    \resizebox{\textwidth}{!}{
        \begin{tabular}{lccccc}
        \toprule
        ~ & Forward & \multicolumn{2}{c}{Backward} & \multicolumn{2}{c}{Update}\\
        \cmidrule(r){2-2} \cmidrule(r){3-4} \cmidrule(r){5-6}
        Methods & A-G(P) & A-G(P) & R-S(G) & R-S(G)/\textbf{A-R(G)} & A-G(P)\\
        \midrule
        ZeRO-1      & 0  
                    & 0 
                    & 0 
                    & \makecell[c]{$\mathbf{2*g*\frac{\Psi}{N}*(N-1)^\dag+}$\\$\mathbf{2*(N-g)*\frac{\Psi}{N}*(N-1)}$} 
                    & \makecell[c]{$g*\frac{\Psi}{N}*(N-1)^\dag+$\\$(N-g)*\frac{\Psi}{N}*(N-1)$} 
                    \\ \specialrule{0em}{0pt}{5pt}
        ZeRO-2      & 0  
                    & 0 
                    & \makecell[c]{$g*s*\frac{\Psi}{N}*(N-1)^\dag+$\\$(N-g)*s*\frac{\Psi}{N}*(N-1)$} 
                    & 0 
                    & \makecell[c]{$g*\frac{\Psi}{N}*(N-1)^\dag+$\\$(N-g)*\frac{\Psi}{N}*(N-1)$} 
                    \\ \specialrule{0em}{0pt}{5pt}
        ZeRO-3      & \makecell[c]{$g*s*\frac{\Psi}{N}*(N-1)^\dag+$\\$(N-g)*s*\frac{\Psi}{N}*(N-1)$} 
                    & \makecell[c]{$g*s*\frac{\Psi}{N}*(N-1)^\dag+$\\$(N-g)*s*\frac{\Psi}{N}*(N-1)$} 
                    & \makecell[c]{$g*s*\frac{\Psi}{N}*(N-1)^\dag+$\\$(N-g)*s*\frac{\Psi}{N}*(N-1)$} 
                    & 0 
                    & 0 
                    \\ \specialrule{0em}{0pt}{5pt}
        MiCS        & $N*s*\frac{\Psi}{M}*(M-1)$ 
                    & $N*s*\frac{\Psi}{M}*(M-1)$ 
                    & $N*s*\frac{\Psi}{M}*(M-1)$ 
                    & \makecell[c]{$\mathbf{2*g*\frac{\Psi}{M}*(g-1)^\dag+}$\\$\mathbf{2*(N-g)*\frac{\Psi}{M}*(g-1)}$}    
                    & 0 
                    \\ \specialrule{0em}{0pt}{5pt}
        ZeRO++      & \makecell[c]{$g*s*\frac{\Psi}{N}*(N-1)^\dag+$\\$(N-g)*s*\frac{\Psi}{N}*(N-1)$} 
                    & $N*s*\frac{\Psi}{M}*(M-1)$ 
                    & \makecell[c]{$g*s*\frac{\Psi}{N}*(N-1)^\dag+$\\$(N-g)*s*\frac{\Psi}{N}*(N-1)$}    
                    & 0   
                    & 0
                    \\ \specialrule{0em}{0pt}{5pt}
        PaRO-IGG    & $N*s*\frac{\Psi}{M}*(M-1)$  
                    & $N*s*\frac{\Psi}{M}*(M-1)$ 
                    & \makecell[c]{$N*s*\frac{\Psi}{N}*(g-1)^\dag+$\\$N*s*\frac{\Psi}{M}*(M-1)$}    
                    & 0   
                    & $N*s*\frac{\Psi}{N}*(g-1)^\dag$ 
                    \\ \specialrule{0em}{0pt}{5pt}
        PaRO-IIG    & $N*s*\frac{\Psi}{M}*(M-1)$ 
                    & $N*s*\frac{\Psi}{M}*(M-1)$ 
                    & $N*s*\frac{\Psi}{M}*(M-1)$    
                    & $N*\frac{\Psi}{N}*(g-1)^\dag$   
                    & $N*\frac{\Psi}{N}*(g-1)^\dag$ 
                    \\ \specialrule{0em}{0pt}{5pt}
        PaRO-NIG    & 0  
                    & 0
                    & $N*\frac{\Psi}{M}*(M-1)$     
                    & $N*\frac{\Psi}{N}*(g-1)^\dag$  
                    & \makecell[c]{$N*\frac{\Psi}{N}*(g-1)^\dag+$\\$N*\frac{\Psi}{M}*(M-1)$}   
                    \\ 
        \bottomrule
        \end{tabular}
    }
    \end{center}
    \vskip -0.1in
\end{table*}

Table ~\ref{tab:mem} shows the single-GPU memory consumption of parameter P, gradient G and optimizer state OS in different solutions.
As can be seen, the single-GPU memory consumption of ZeRO-1, ZeRO-2, and ZeRO-3 is not affected by the number of groups, as they only perform global sharding operations.
MiCS shards the entire model state within the group and introduces inter-group redundancy. 
As a result, the memory of MiCS linearly increases with the number of groups.
Based on ZeRO-3, ZeRO++ additionally retains the intra-group sharding of model parameters, while PaRO-IGG only retains the intra-group sharding of model parameters.
Therefore, the memory of PaRO-IGG and ZeRO++ slowly increases with the number of groups.
PaRO-IIG shards model parameters and gradients within groups, further increasing memory redundancy.
Based on ZeRO-2, PaRO-NIG shards gradient groups, and its memory also increases slowly as the number of groups increases.

Table ~\ref{tab:com} shows the total communication volume of different solutions in the Forward, Backward and Update stages with a mini-batch input.
As can be seen from Table ~\ref{tab:com}, each GPU in ZeRO-1 performs gradient accumulation locally, and only performs a global synchronization after gradient accumulation.
MiCS performs intra-node communication in the Forward and Backward stages, and only performs a partial gradient all-reduce operation for parameter update.
For ZeRO++, due to the secondary intra-node sharding of the collected model parameters in the Forward stage, the parameters can be collected using an intra-node all-gather operation in the Backward stage.
Since MiCS, PaRO-IGG and PaRO-IIG shard the model parameters within the group, the all-gather operation in forward and backward computation is intra-group communication.
Compared with ZeRO-3, these solutions increase the size of a single transmission ($\frac{\Psi}{M}$ vs. $\frac{\Psi}{N}$) and reduce the number of communications ($s*(M-1)$ vs. $s*(N-1)$), which can improve the bandwidth utilization within the group.
Compared with ZeRO-2, PaRO-NIG splits the global reduce-scatter of the gradient into two steps: intra- and inter-group reduce-scatters.

\begin{figure}[t]
    \centering
    \includegraphics[width=0.40\textwidth]{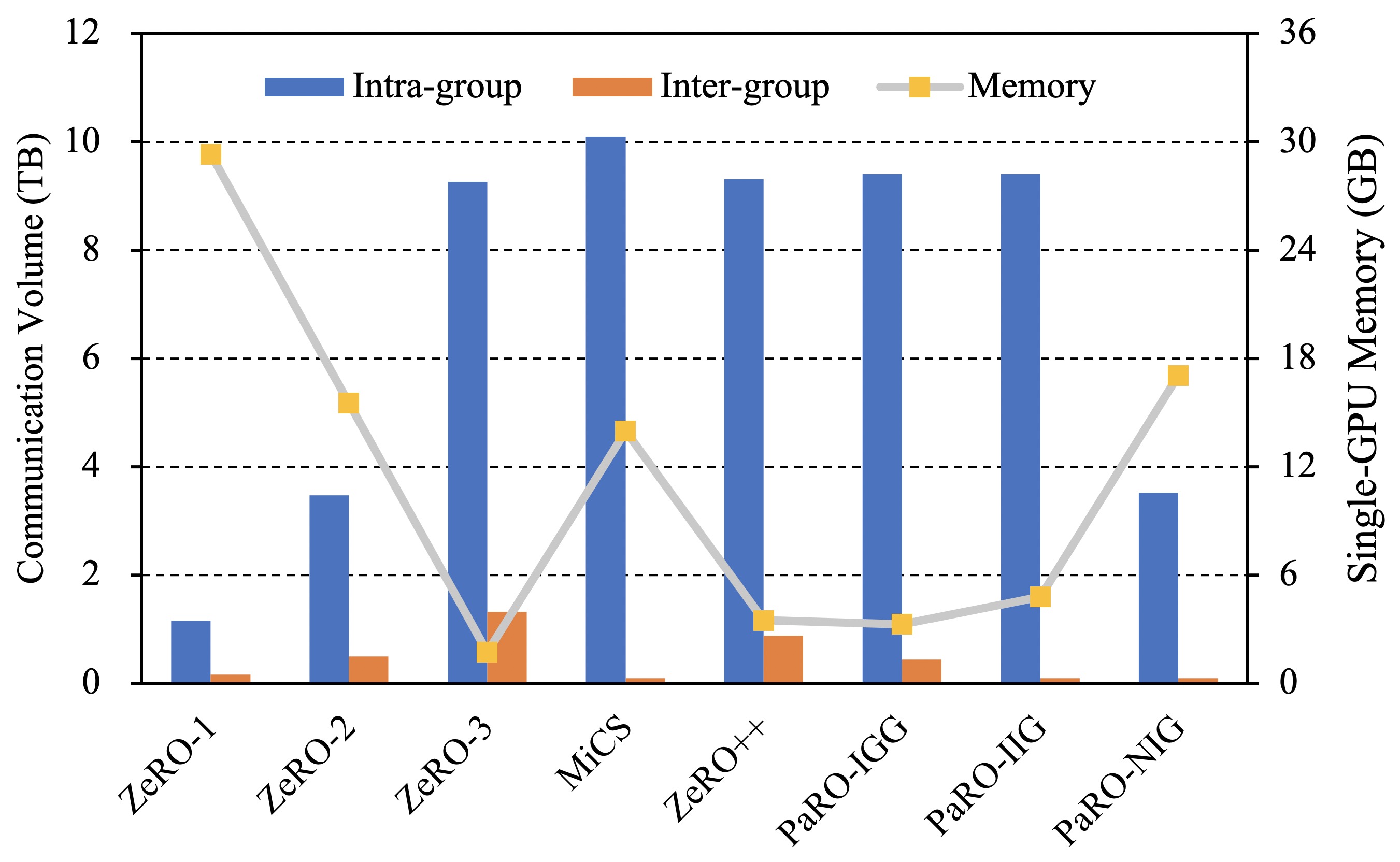}
    \caption{Under the conditions of $\Psi=7B, N=64, s=8, g=8$, the total communication volume (intra- and inter-group), and single GPU memory of model states.
    }
    \label{fig:mem-com}
\end{figure}
Figure ~\ref{fig:mem-com} shows the total intra- and inter-group communication volume of different solutions under the condition of $\Psi=7B, N=64, s=8, g=8$, as well as the memory occupation of a single GPU. 
ZeRO-1, ZeRO-2, and ZeRO-3 progressively shard the model state, resulting in a near-linear reduction in single-GPU memory and a near-linear increase in intra- and inter-group communication volume.
Compared with ZeRO-3, the single-GPU memory of ZeRO++, PaRO-IGG and PaRO-IIG increases slightly, while that of MiCS increases significantly;
the intra-group communication volume of ZeRO++, PaRO-IGG and PaRO-IIG is the same as ZeRO-3, only slightly increased in MiCS;
the inter-group communication volume of MiCS, ZeRO++, PaRO-IGG and PaRO-IIG is smaller than ZeRO-3, with MiCS and PaRO-IIG being the lowest.
Compared with ZeRO-2, the intra-group communication volume and single-GPU memory of PaRO-NIG increase slightly, while the inter-group communication volume decreases significantly.
In summary, PaRO offer a better balance between memory consumption and communication costs for LLM training.

\subsection{Use PaRO in complex ML systems}
PaRO can also be applied in complex ML systems. For instance, in the PPO step of RLHF, it is sometimes necessary to deploy partial states of multiple models, such as Actor, Critical, and Reward models, on a single GPU. Each model has different memory and communication requirements. By applying different PaRO strategies to different models, it is possible to better balance the cost of each model, thereby improving the end-to-end PPO speed.

\section{Experiments and Analysis}
In this section, we perform end-to-end training to evaluate the throughput and scalability of the proposed PaRO.
Afterwards, we evaluate the transmission efficiency of the HO-Ring communication topology.
Finally, we demonstrate the consistent convergence of PaRO and ZeRO, which validates the correctness of our system. 

\subsection{Experiment Environments}
Our experimental cluster consists of up to 16 DGX nodes, with each node containing 8 Ampere A100 SXM3 80GB GPUs. 
The GPUs in each node are interconnected via NVLink/NVSwitch with a bidirectional bandwidth of up to 600GB/s.
These nodes are connected through 8 InfiniBand adapters without NVIDIA SHARP, which can provide more than 100GB/s of inter-node bandwidth. 
The software environment includes CUDA-11.7, DeepSpeed-v0.10.0, PyTorch-v1.9.2, and NCCL-v2.14.3.

\subsection{Throughput Performance}
We used ZeRO-2 and ZeRO-3 in Deepspeed as baselines to implement PaROs with different sharding strategies.
To evaluate the performance of PaROs, we compared them with current state-of-the-art solutions, including: ZeRO, ZeRO-3, MiCS, ZeRO++ and FSDP-hs.
ZeRO-1 was not considered due to its inability to run the smallest scale model in our experiments.
We used two LLMs with different parameter sizes: LLaMA-7B and LLaMA-65B \cite{llama2023}, to evaluate the throughput and acceleration performance at varying GPU counts.
For the LLaMA-65B model, we activated checkpointing to ensure successful training.
The C4 corpus in RedPajama was used as the training data set.
During training, we set the sequence length to 512, the batch size to 40 (divided into 4 micro-batches), the number of gradient accumulation steps to 10, and mixed precision.
All throughput data reported was the average of 100 iterations.

\begin{figure*}[t]
    \centering
    \subfigure[Throughput of LLaMA-7B.]{
        \begin{minipage}[b]{0.3\textwidth}
        \includegraphics[width=1\textwidth]{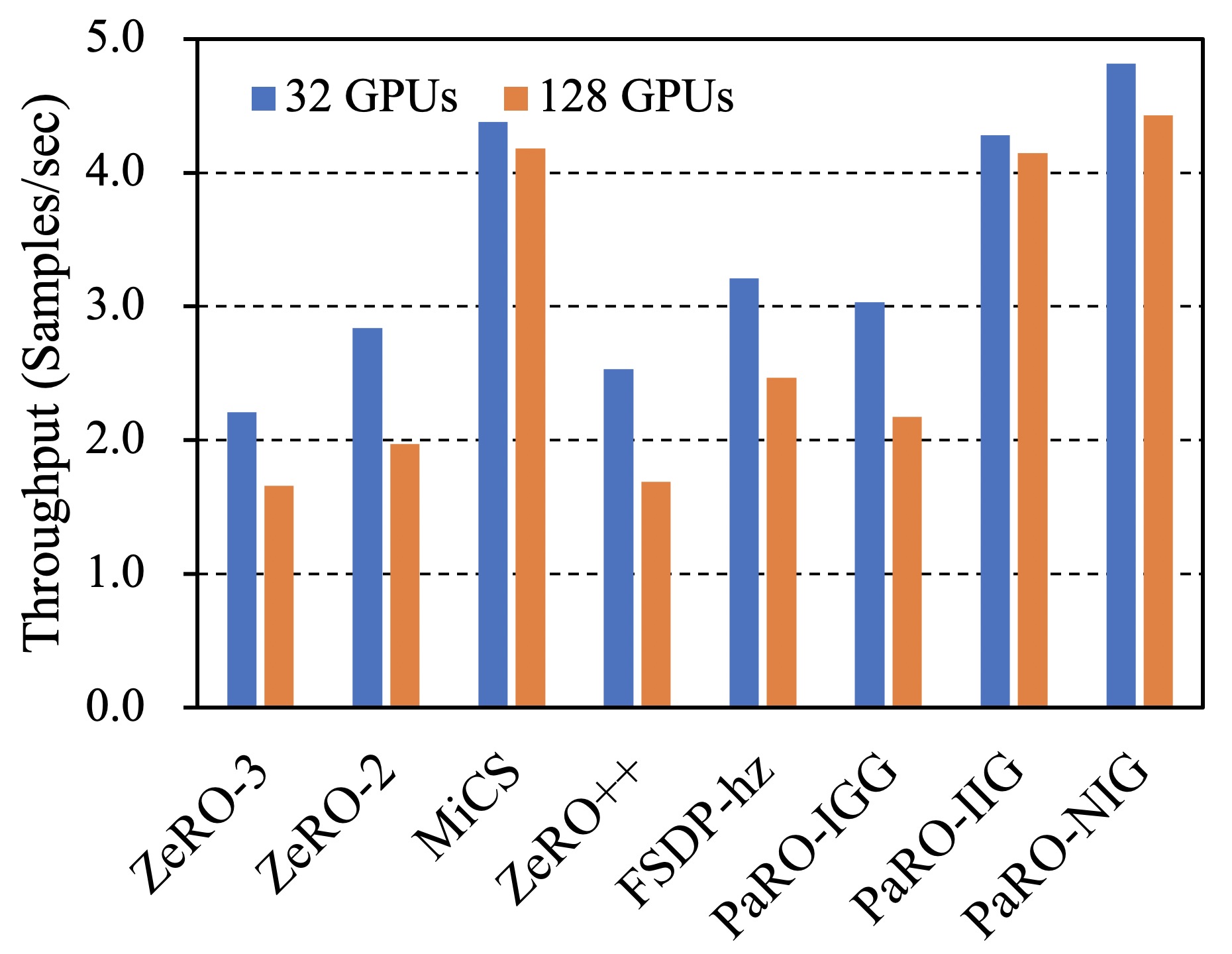}
        \end{minipage}
        \label{fig:Thro-7B}
    }
    \subfigure[Peak memory of LLaMA-7B.]{
        \begin{minipage}[b]{0.3\textwidth}
        \includegraphics[width=1\textwidth]{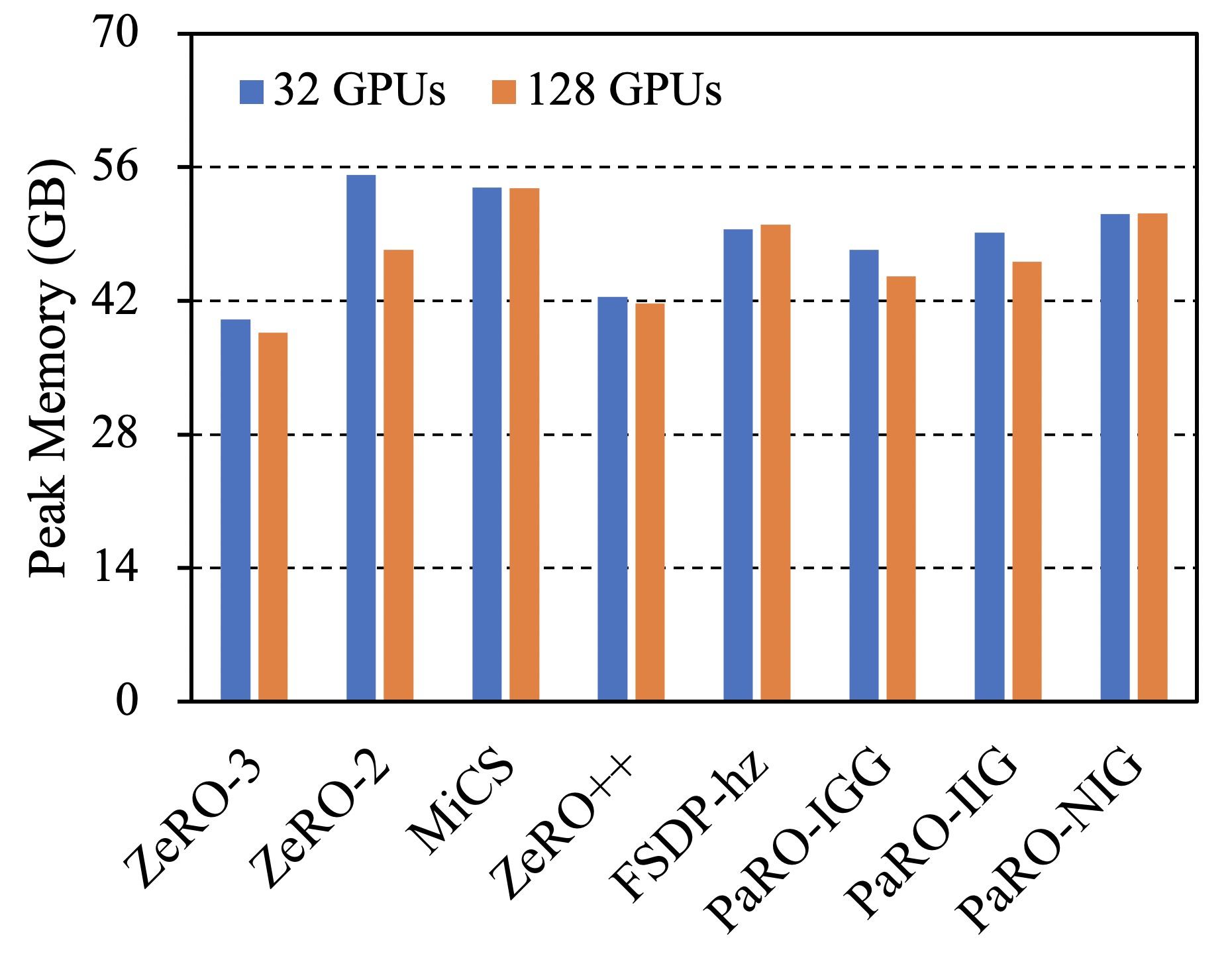}
        \end{minipage}
        \label{fig:peakmem}
    }
    \subfigure[Throughput of LLaMA-65B.]{
        \begin{minipage}[b]{0.3\textwidth}
        \includegraphics[width=1\textwidth]{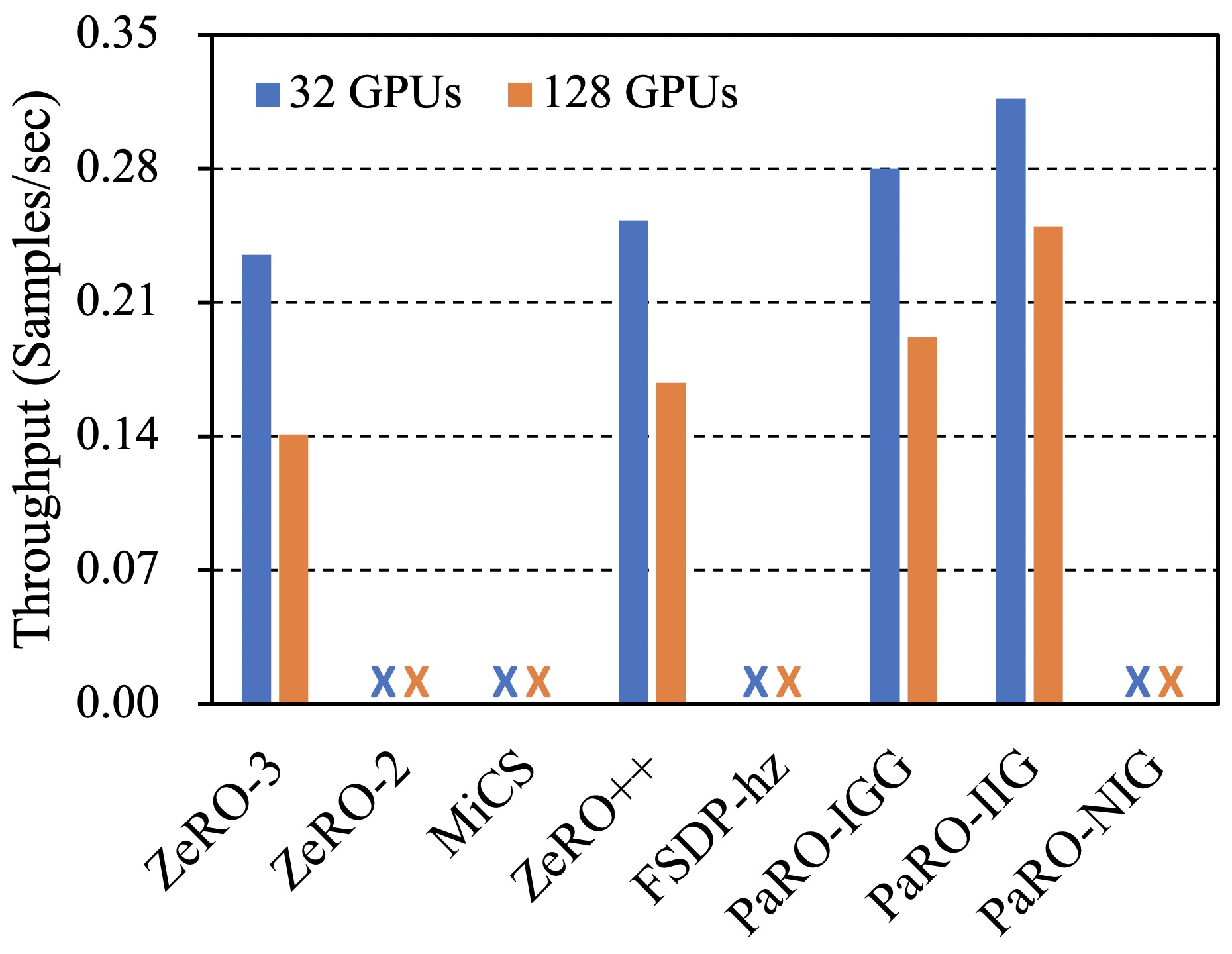}
        \end{minipage}
        \label{fig:Thro-65B}
    }
    \caption{Throughput and peak memory of LLaMA-7B and LLaMA-65B models in different solutions. The cross indicates OOM.}
    \label{fig:Thro_Mem}
\end{figure*}

\begin{figure*}[t]
    \centering
    \begin{minipage}[t]{0.3\linewidth}
        \centering
        \includegraphics[width=0.9\textwidth]{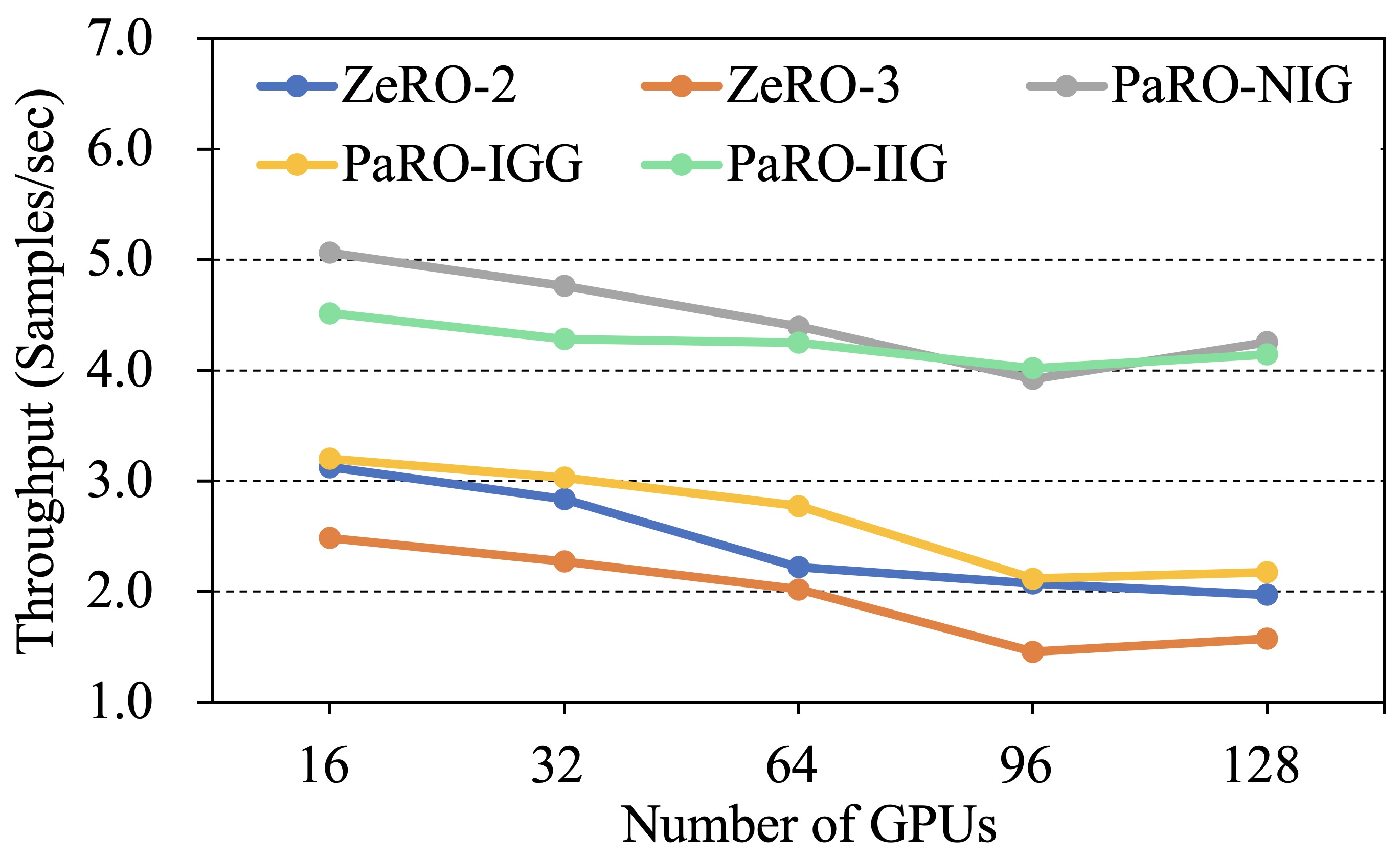}
        \caption{Throughput with different number of GPUs.}
        \label{fig:th-com}
    \end{minipage}
    \quad
    \begin{minipage}[t]{0.3\linewidth}
        \centering
        \includegraphics[width=0.8\textwidth]{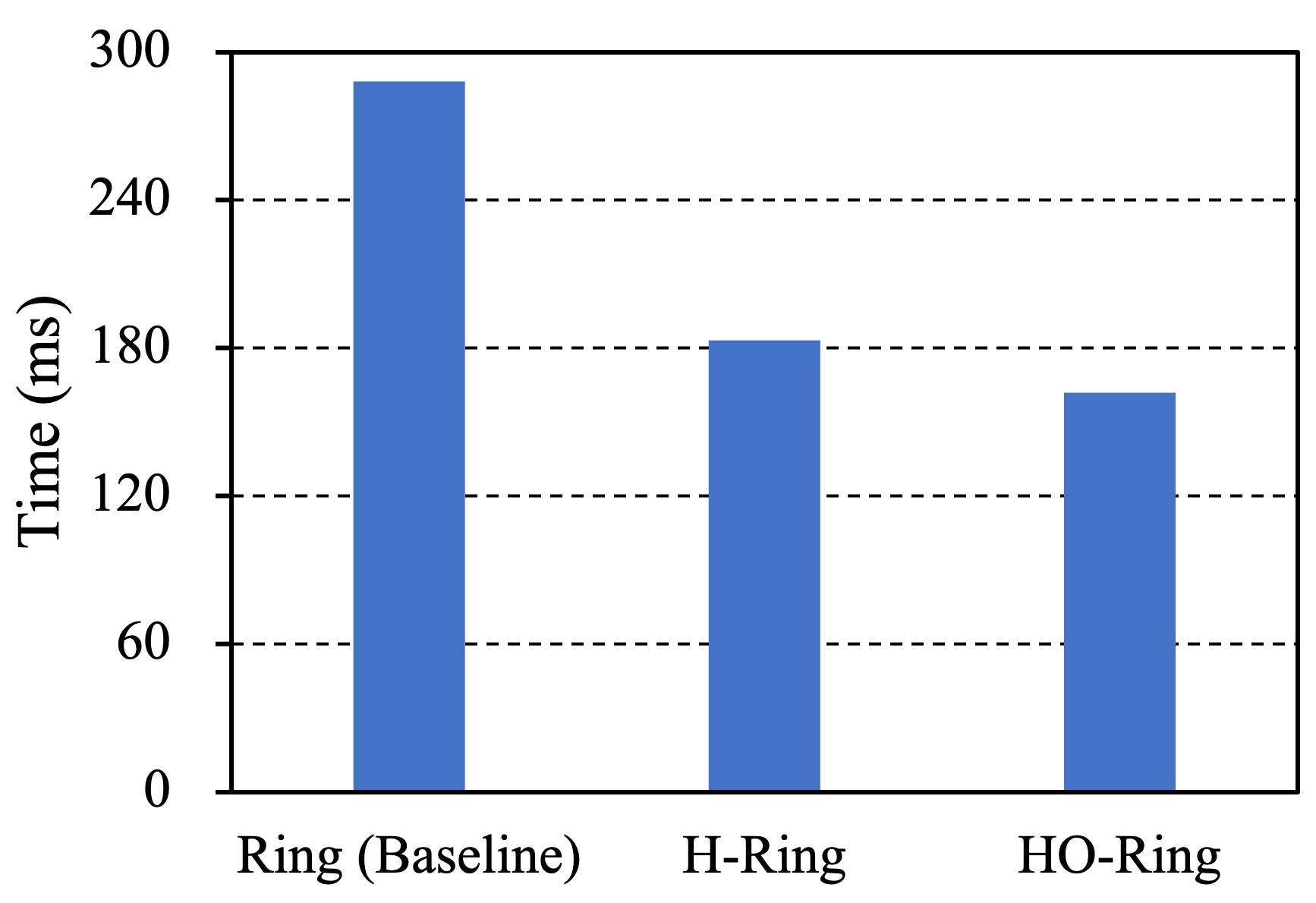}
        \caption{Time comparison of different strategies for all-gather operation.}
        \label{fig:ring_com}
    \end{minipage}
    \quad
    \begin{minipage}[t]{0.3\linewidth}
        \centering
        \includegraphics[width=0.95\textwidth]{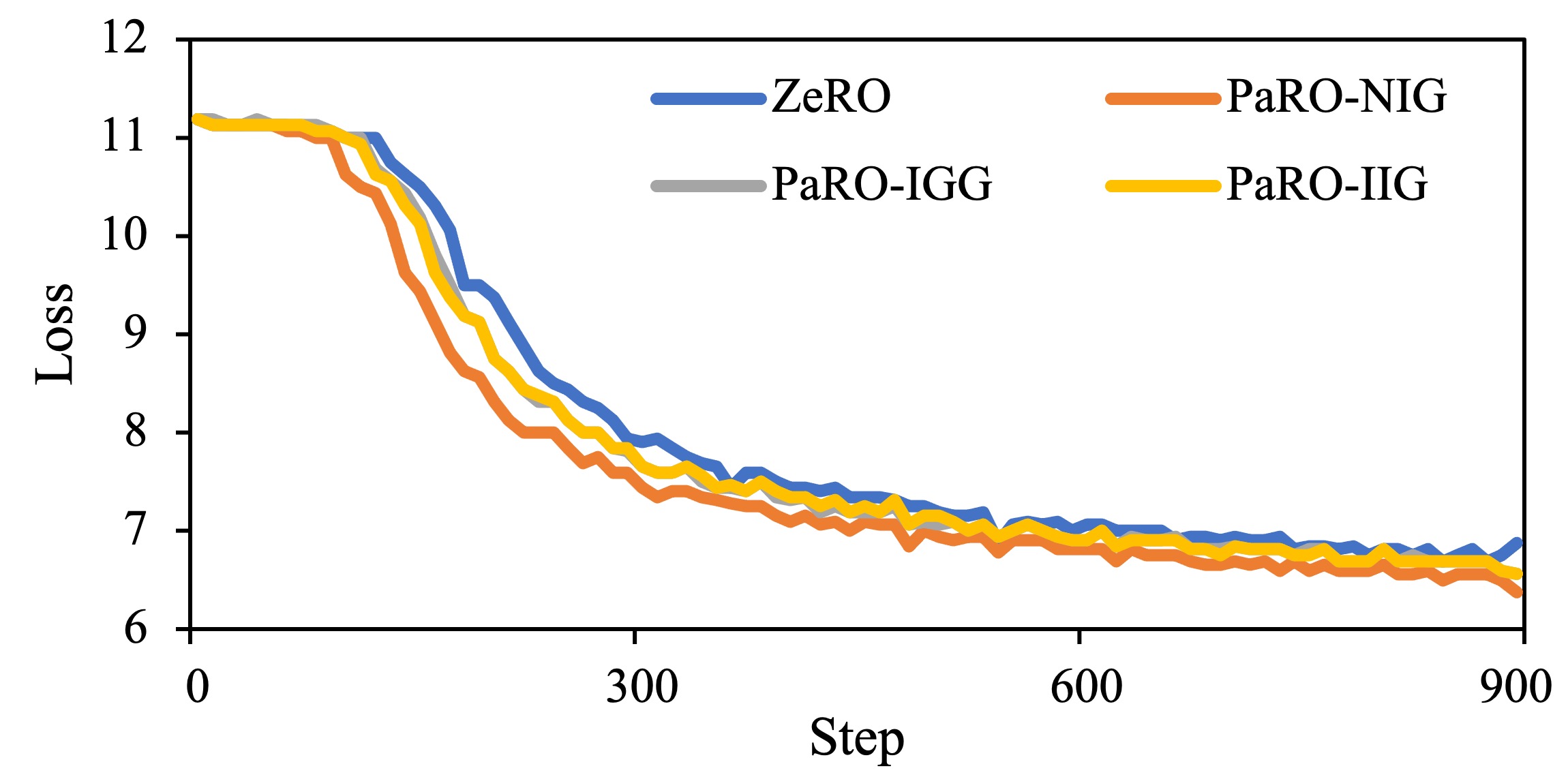}
        \caption{Training convergence for LLaMA-7B.}
        \label{fig:loss}
    \end{minipage}
\end{figure*}

Figure~\ref{fig:Thro_Mem} shows the throughput and peak memory of LLaMA-7B and LLaMA-65B in different solutions.
In Figure~\ref{fig:Thro-7B}, the throughput of PaRO-IGG is only better than that of ZeRO-3 and ZeRO++; the throughput of PaRO-IIG is almost the same as that of MiCS.
Compared with the baseline ZeRO-3, the throughput of PaRO-IGG and PaRO-IIG is improved by 1.37x and 1.94x (with 32 GPUs), 1.31x and 2.50x (with 128 GPUs), respectively.
Compared with the baseline ZeRO-2, the throughput of PaRO-NIG is improved by 1.70x (with 32 GPUs) and 2.25x (with 128 GPUs), respectively.
For the small-scale LLaMA-7B, PaRO-NIG approaches show higher throughput in clusters with 32 and 128 GPUs.

Figure ~\ref{fig:peakmem} shows the maximum reserved memory during training for different solutions. 
The peak memory of ZeRO-3 is the smallest.
Since both MiCS and FSDP-hs adopt an intra-group sharding strategy, their peak memory is only related to the number of GPUs in the group, but not to the number of GPUs in the cluster. 
The peak memory of PaRO-IGG, PaRO-IIG and PaRO-NIG increase slightly compared to baseline ZeRO-3 and ZeRO-2 respectively.
The peak memory of all three PaROs is smaller than MiCS.

Figure~\ref{fig:Thro-65B} presents the single GPU throughput of LLaMA-65B with different approaches. 
Since LLaMA-65B requires finer-grained sharding, only ZeRO-3, ZeRO++, PaRO-IGG, and PaRO-IIG can perform training, while other solutions suffer from out-of-memory (OOM) issues.
Compared with the baseline ZeRO-3, the throughput of PaRO-IGG and PaRO-IIG is improved by 1.19x and 1.35x (with 32 GPUs), 1.36x and 1.77x (with 128 GPUs), respectively.
For the large-scale LLaMA-65B, training efficiency is higher with the PaRO-IGG and PaRO-IIG compared to ZeRO-3.

\subsection{Near-linear Scalability}
To analyze the relationship between throughput and GPU resources, we collected the single-GPU throughput of PaRO and ZeRO under different GPU numbers, as shown in Figure ~\ref{fig:th-com}. 
The experiments were conducted using LLaMA-7B.
Overall, under the same GPU resource conditions, the single-GPU throughput of PaRO is higher than the baseline ZeRO-2 and ZeRO-3. 
The single-GPU throughput of different approaches gradually decreases as the number of GPUs increases.
The throughput of PaRO-IIG decreases the least, and the throughput of ZeRO-2 decreases the most.
Since NCCL adopts a multi-machine communication method based on Double Binary Tree ~\cite{SANDERS2009581}, the communication efficiency of 96 GPUs (not an integer power of 2) is lower than that of 64 and 128 GPUs.
It can be seen that as the cluster size increases, PaRO-IIG can maintain near-linear scalability.

\subsection{HO-Ring Communication Performance}
In the section, we performed experiments using 16 DGX nodes, with a total communication volume set to 1GB. 
We measured the communication time of the all-gather operation with the traditional Ring (baseline), H-Ring and HO-Ring.
The communication times of traditional Ring, H-Ring and HO-Ring are 288ms, 183ms and 162ms respectively. 
Compared with Ring and H-Ring, the communication time of HO-Ring is reduced by 36.5\% and 11.5\% respectively. 
Therefore, HO-Ring can significantly improve communication efficiency by improving the communication topology.

\subsection{Model Convergence}
We used LLaMA-7B and C4 corpus in RedPajama to evaluate the convergence of PaRO.
During training, we set the sequence length to 128, the batch size to 1024 (divided into 8 micro-batches) and the number of gradient accumulation steps to 8.
The loss validation process does not aim to produce exactly the same loss as ZeRO but to ensure the convergence behaviours are the same. 
As shown in Figure ~\ref{fig:loss}, PaRO provides the same convergence as ZeRO.

\section{Conclusion}
In this paper, we present PaRO, a system balances the memory occupation and communication costs across diverse training scenarios. PaRO provides comprehensive options which reduces the communication cost of inter-group communication with minor memory redundancy by fine-grained sharding strategy, thereby improving the training efficiency. Additionally, we propose a HO-Ring communication topology to enhance collective communication efficiency between nodes or across switches. 
We evaluate PaRO on different training workloads on large-scale clusters. 
PaRO outperforms ZeRO by up to 2.50$\times$ and demonstrates near-linear scalability in various industrial-level training settings.


\bibliography{main}
\bibliographystyle{mlsys2024}



\end{document}